\documentclass{article}
\usepackage{spconf,amsmath,graphicx}
\usepackage{adjustbox,booktabs,multirow,multicol,xcolor,amsfonts,amssymb,url}
\usepackage{tikz,enumitem,titlesec}
\usepackage{comment}
\definecolor{DarkRed}{rgb}{0.7, 0.0, 0.0}
\definecolor{DarkGreen}{rgb}{0.0, 0.3, 0.0}

\title{PawPrint: Whose Footprints Are These?\\Identifying Animal Individuals by Their Footprints}
\name{Inpyo Song$^{\star}$, Hyemin Hwang, Jangwon Lee$^{\dagger}$\thanks{$^{\star}$ songinpyo@skku.edu, $^{\dagger}$ Corresponding author: leejang@skku.edu}}
\address{Department of Immersive Media Engineering, Sungkyunkwan University, South Korea}

\begin{document}
%
\maketitle

\begin{abstract}
In the United States, as of 2023, pet ownership has reached 66\% of households and continues to rise annually.
This trend underscores the critical need for effective pet identification and monitoring methods,
particularly as nearly 10 million cats and dogs are reported stolen or lost each year.
However, traditional methods for finding lost animals like GPS tags or ID photos have limitations-they can be removed,
face signal issues, and depend on someone finding and reporting the pet.
To address these limitations, we introduce \textsc{PawPrint} and \textsc{PawPrint+}, the first publicly available datasets focused on \emph{individual}-level footprint identification for dogs and cats.
Through comprehensive benchmarking of both modern deep neural networks (e.g., CNN, Transformers) and classical local features, we observe varying advantages and drawbacks depending on substrate complexity and data availability.
These insights suggest future directions for combining learned global representations with local descriptors to enhance reliability across diverse, real-world conditions.  
As this approach provides a non-invasive alternative to traditional ID tags,
we anticipate promising applications in ethical pet management and wildlife conservation efforts.

\end{abstract}


\section{Introduction}
\label{sec:intro}
As of 2023, 66\% of households in the United States, approximately 86.9 million homes own pets,
with the number of pet owners increasing annually \cite{APPA2023pet}.
This growing trend is accompanied by significant challenges:
each year, nearly 10 million cats and dogs are reported stolen or lost,
and approximately 6.3 million companion animals enter animal shelters \cite{rodriguez2022trends}.
The loss of a pet can profoundly affect pet owners, as pets are often cherished as beloved family members.
Furthermore, these lost pets contribute to a range of issues,
including animal welfare concerns such as euthanasia \cite{marston2004happens},
public safety risks such as bites and traffic accidents involving animals \cite{canal2018dogs},
and broader social and ecological impacts \cite{bernete2021amelioration}.

\begin{figure}[!t]
\centering
\includegraphics[width=\columnwidth]{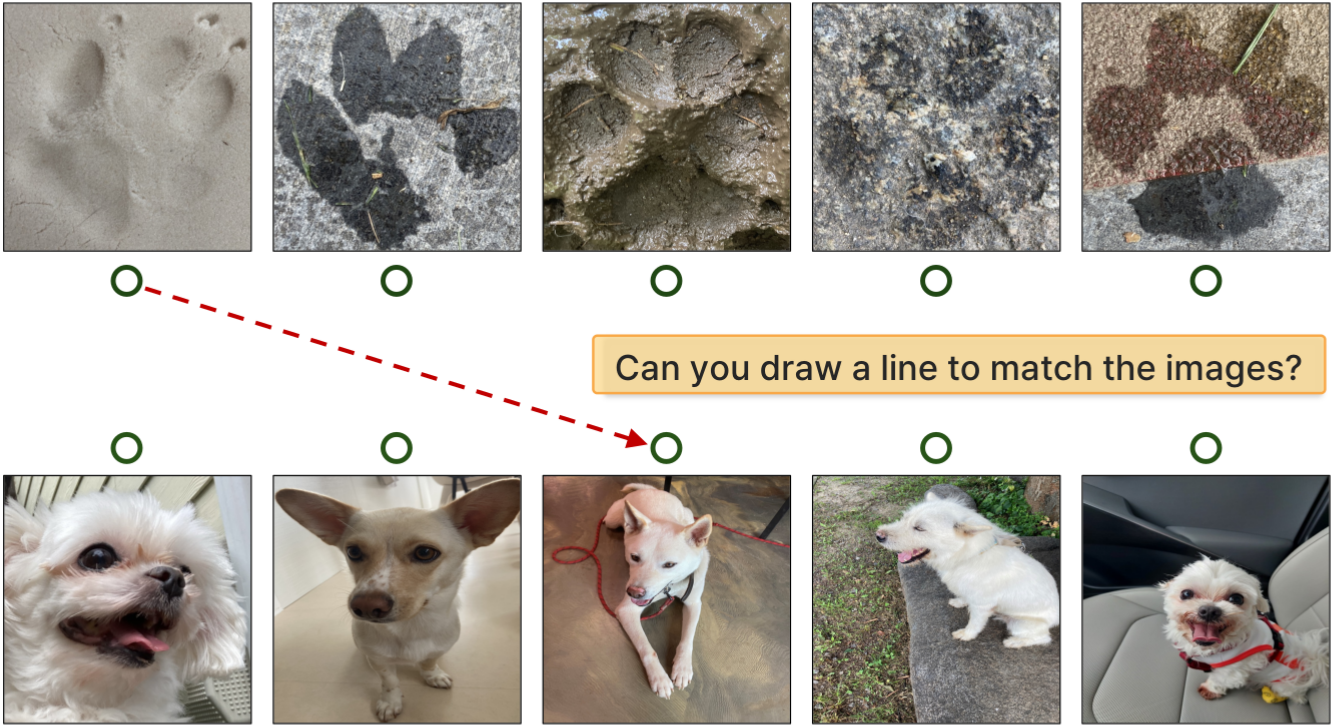}
\caption{
Can you correctly match each print to its corresponding individual animals? 
In this work, we explore a non-invasive, footprint-based identification framework 
that could benefit both pet owners and wildlife conservation efforts.
}
\vspace{-1em}
\label{fig:front_image}
\end{figure}

Traditionally, people rely on owner-provided photos or personalized ID tags to locate their missing pets.
However, these methods depend on someone spotting the animal, which is not always the case.
Although GPS-based tags with monitoring systems are now widely used,
their effectiveness can be compromised if the tags are intentionally removed or if the animals are in GPS signal weak spots,
such as indoors or in dense foliage.
Animal biometric systems, like recognition chips,
are also prevalent for monitoring wildlife and managing endangered species. 
However, researchers increasingly seek less invasive animal identification methods
in wildlife research and management due to the potential disturbance of normal animal behavior by attached sensors.
These sensors also pose risks of physical discomfort or injury, challenges with sensor removal and data loss,
logistical issues with battery life and maintenance, complex data interpretation, ethical considerations, and high costs.

\begin{figure*}[!t]
  \centering
  \includegraphics[width=\linewidth]{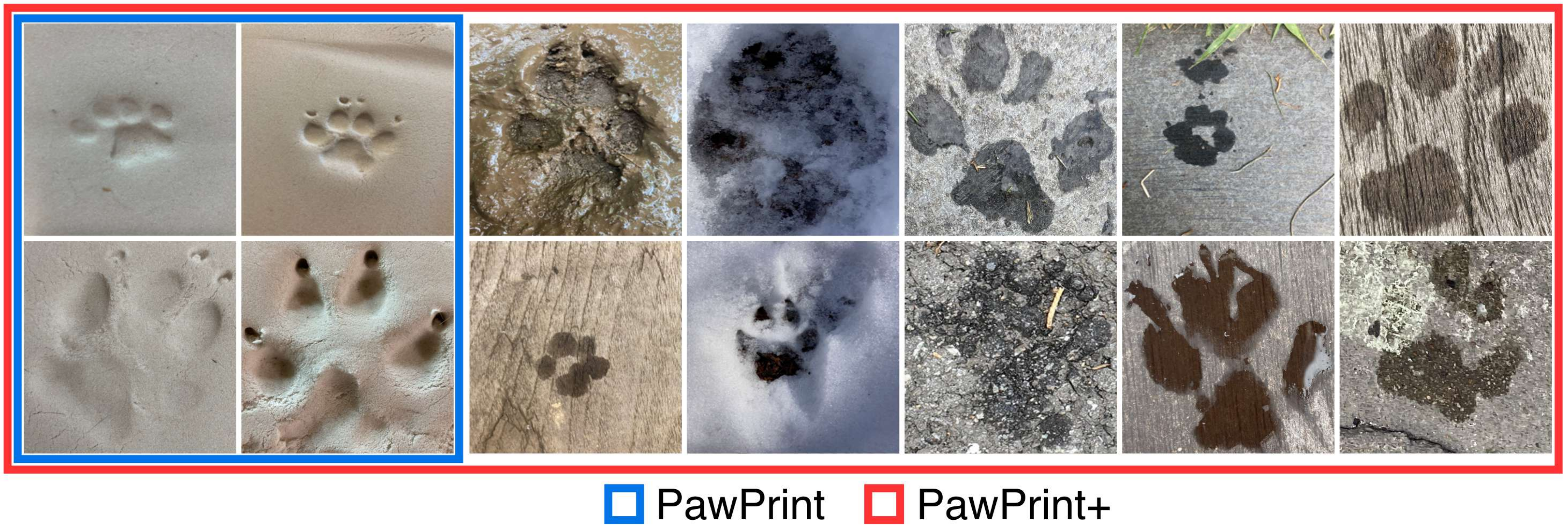}
  \caption{
    The proposed \textsc{PawPrint} dataset, with two versions:
    \textsc{PawPrint} (sand/clay only) and\textsc{PawPrint+} (additional natural terrains like wood, dirt, snow, rock, asphalt).
    Our goal is to facilitate individual-level animal identification using their footprints.
  }
  \vspace*{-1em}
   \label{fig:pawprint_dataset}
\end{figure*}

Consequently, the growing importance of non-invasive methods in wildlife health and disease research
is a response to these limitations
\cite{schilling2022review}.
These methods are recognized for their efficacy in identifying individual animals more effectively than intrusive techniques like trapping,
as well as for being cost-effective and aligning better with animal welfare.
Among these, Footprint-based Identification Technology (FIT)
\cite{li2018using}
stands out as one of the most promising non-invasive methods for wildlife tracking.
While effective, the FIT process can be complex and time-consuming,
requiring detailed pre-processing of footprint feature points and sizes.

Therefore, this study introduces the first publicly available datasets dedicated to individual-level footprint identification 
for dogs and cats, which do not require any specialized pre- or post-processing. 
Specifically, we introduce two new datasets, \textsc{PawPrint} and \textsc{PawPrint+}, covering diverse dog and cat footprints.
\textsc{PawPrint} contains relatively clear imprints from controlled substrates (sand, clay), 
providing a straightforward testbed to assess end-to-end learning of unique pawprint features. 
Building on this, \textsc{PawPrint+} extends to footprints captured in varied natural environments such as mud, asphalt, stone, and wood, 
offering a more realistic and challenging setting. 
In practical scenarios, such variations are critical for identifying missing or stray animals whose pawprints may not be found in ideal conditions.

We comprehensively benchmark both modern deep learning architectures and classical local feature methods 
on \textsc{PawPrint} and \textsc{PawPrint+}, demonstrating that footprint-based identification is indeed feasible 
in both controlled and more challenging real-world settings. 
These experiments illuminate the benefits and drawbacks of each approach, 
suggesting directions for hybrid techniques that harness the strengths of deep learning representations 
and robust local descriptors. 
By introducing publicly accessible datasets and evaluating a range of methods, we hope that this paper can contribute to a promising approach for ethical and efficient animal tracking,
with significant potential applications in both pet management and wildlife conservation.



\section{Related Work}\label{sec:related work}
\noindent \textbf{Visual Biometrics for Animal Identification.}
Accurate identification of individual animals is a fundamental task in wildlife management and conservation, especially for monitoring populations of stray or wild species. Traditional visual biometric approaches have focused on unique coat or skin patterning—e.g., tigers \cite{hiby2009tiger}, polar bears \cite{anderson2010computer}, zebras \cite{lahiri2011biometric}, and dog noseprint patterns \cite{bae2021dog}.
While successful in controlled settings, these methods often require capturing explicit body patterns or specific viewpoints, posing significant hurdles when dealing with free-roaming or elusive animals \cite{vcermak2024wildlifedatasets,shinoda2025petface}.

\noindent \textbf{Footprint-Based Identification (FIT).}
In response to the limitations of body-pattern identification, footprint-based identification technology (FIT) has emerged as a non-invasive alternative \cite{li2018using,alibhai2008footprint}.
FIT exploits the morphology of pawprints, which can be collected from natural terrains, thus reducing stress on animals. However, these approaches traditionally rely on manual landmark annotations and expert-driven feature engineering—processes that are both time-consuming and prone to human error.
For instance, \cite{kistner2022s} manually annotated 11 landmarks to classify southeast asian otter species from footprints, while \cite{li2018using} identified sex and age classes in giant pandas using 7 annotated footprint landmarks.

\noindent\textbf{Deep Learning for Footprint Recognition.}
Advances in deep neural networks have mitigated the need for manual feature design.
For instance, Kistner et al. \cite{kistner2023can} demonstrated automated species classification of southeast asian otter footprints without explicit landmarking.
Similarly, OpenAnimalTracks \cite{shinoda2024openanimaltracks} explored convolutional and transformer-based architectures (e.g., ResNet, VGG, ViT) for multi-species footprint recognition with minimal human intervention.
Despite these gains, most existing works focus on species-level identification and lack of publicly available datasets suitable for individual-level footprint classification.

Building on these developments, our contribution is a first publicly accessible dataset to best of our knowledge, which dedicated to identifying individual animals specifically for cats and dogs using their footprints.
By combining footprints captured in both controlled and natural environments, our dataset aims to address the scalability and generalization challenges, while eliminating the need for extensive manual annotations through modern deep learning approaches.

\section{PawPrint Datasets}
Despite growing interest in non-invasive animal identification technologies, publicly accessible footprint datasets remain scarce.
Existing collections are typically proprietary or limited to wildlife monitoring scenarios, hindering broader research and practical adoption.
To address this gap, we present \textsc{PawPrint}, a dataset of dog and cat footprints, and its extended version, \textsc{PawPrint+}.
To our knowledge, this is the first publicly available dataset for individual-level footprint identification of these two common pets,
aiming to stimulate new research in cost-effective, non-invasive tracking.


\begin{figure}[!t]
  \includegraphics[width=\linewidth]{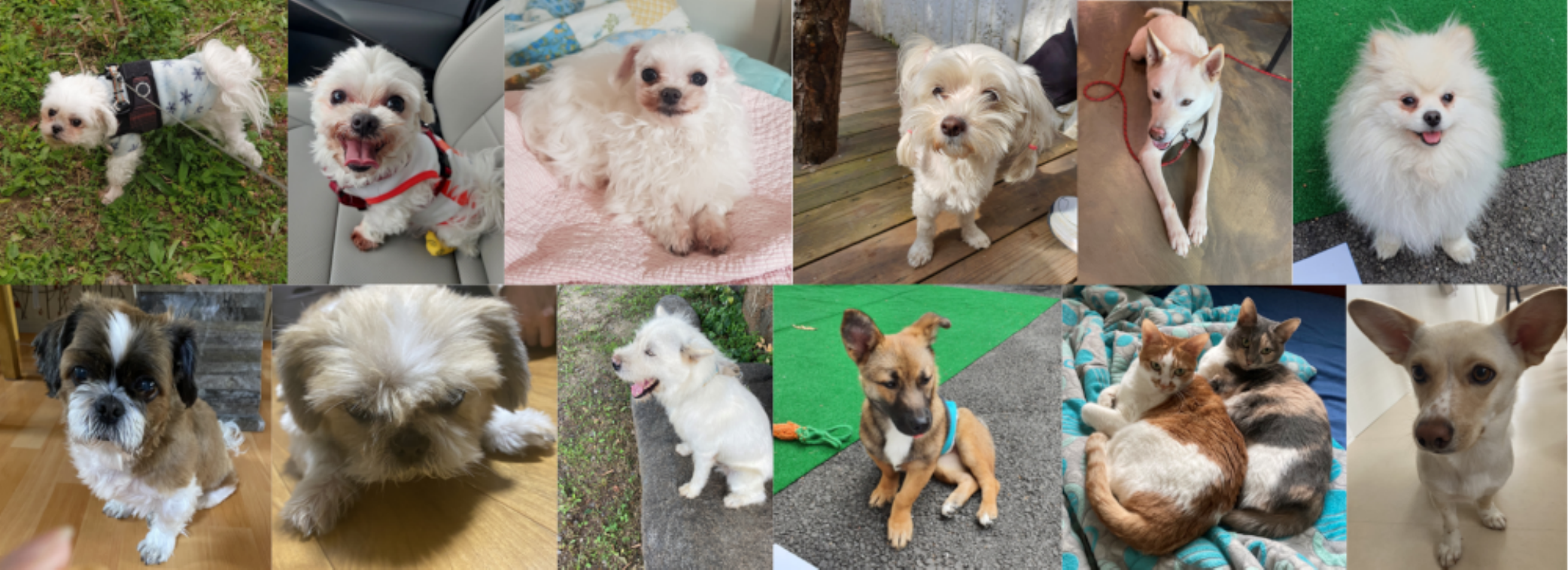}
  \caption{
  Sample pictures of dogs and cats from the \textsc{PawPrint} dataset. 
  In total, we collected 2{,}020 footprint images from 13 dogs and 7 cats across both \textsc{PawPrint} and \textsc{PawPrint+}.
  }
   \label{fig:animal-pictures}
\end{figure}

\subsection{Dataset Statistics}
The initial dataset, \textsc{PawPrint}, consists of 933 footprints collected from
13 dogs and 7 cats (327 cat footprints; 606 dog footprints).
All prints were obtained in controlled, low-clutter substrates (sand, clay) to yield high-contrast imprints.
This controlled setting makes \textsc{PawPrint} ideally suited for validating that end-to-end deep learning models
can reliably extract unique pawprint features for individual identification.

To test generalization under diverse terrain conditions, we also developed \textsc{PawPrint+},
which comprises 1,662 footprints from 12 dogs.
This dataset includes footprints captured on seven naturally varying surfaces—wood, dirt, rock,
asphalt, snow, sand, and clay—to reflect real-world challenges such as partial or noisy imprints.
Specifically, the distribution spans 495 sand, 492 wood, 486 asphalt, 85 clay, 77 rock,
23 dust, and 4 snow footprints.
This broader diversity ensures a more robust training ground for algorithms targeting field-deployable solutions.

\begin{figure}[!t]
  \includegraphics[width=\linewidth]{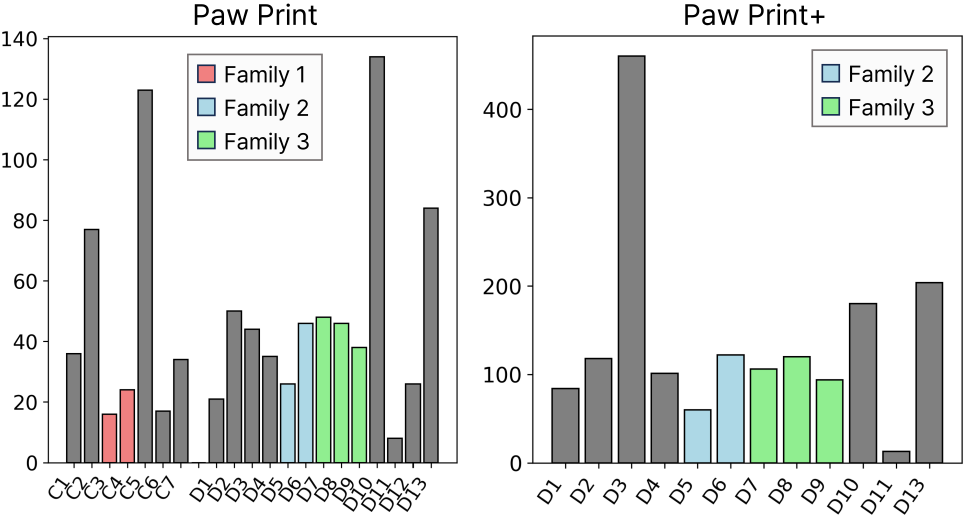}
  \caption{
  \textsc{PawPrint} and \textsc{PawPrint+} dataset distribution.
  }
  \vspace{-1.5em}
   \label{fig:dataset distribution}
\end{figure}

\paragraph{Train/Test Split.}
To enable balanced evaluation, we divide \textsc{PawPrint} into training and testing sets 
using a 7:3 ratio while preserving the proportion of cat and dog footprints (Figure~\ref{fig:dataset distribution}).
\textsc{PawPrint+} uses footprints on sand and clay for training, while footprints on other natural surfaces (wood, dirt, rock, asphalt, snow) are used for testing.
Specifically, this approach allocates 580 images for training 
and 1{,}082 images for testing, mirroring a realistic scenario in which 
reference pawprints are gathered under controlled conditions, 
but actual identification must occur under more diverse and challenging environments.


\subsection{Footprint Image Collection}
The collection of footprint images was conducted with careful consideration for the animals' wellbeing and natural behavior.
For dogs, we captured footprints in various outdoor environments to ensure diversity and real-world applicability.
However, the process presented unique challenges for cats, as they are predominantly indoor animals. To respect their natural habits and avoid invasive practices, cat footprints were collected in indoor or controlled environments.
Ethical considerations were paramount throughout the data collection process.
All images were captured with the full consent and under the supervision of the pets' owners, aligning with findings from Girault et al. \cite{girault2022dog}, which emphasize the positive influence of an owner's presence on a pet's sense of security during examinations.
Recognizing that footprints can vary based on the animal's weight and gait, we aimed for a realistic and varied data collection.
Instead of manually obtaining footprints, we encouraged natural movement, such as walking or running, to capture authentic traces.
These footprints were then photographed, with each image annotated with bounding boxes and an animal ID.

\subsection{Dataset Composition}
Both \textsc{PawPrint} and \textsc{PawPrint+} feature primarily mixed-breed cats and dogs, captured from 13 dogs and 7 cats across both datasets.
An interesting aspect is the inclusion of three sets of closely related animals
(two dog family groups and one family cat pair), enabling the study of fine-grained intra-species variability.
This familial overlap tests the ability of identification methods to differentiate genetically similar individuals
who may share subtle morphological traits in their pawprints.

\begin{table}[t]
    \centering
    \resizebox{\columnwidth}{!}{%
    \begin{tabular}{lcccc}
        \toprule
        \textbf{Method} & \textbf{Accuracy} & \textbf{Recall} & \textbf{Precision} & \textbf{F1-score} \\
        \midrule
        \multicolumn{5}{l}{\emph{Linear Probing}} \\
        \midrule
        ResNet-50 &  79.31  &  77.65  &  79.56  &  77.25  \\
        VGG-16    &  65.52  &  66.13  &  67.97  &  62.52  \\
        Eff.-B1 &  74.83  &  73.88  &  74.23  &  72.90  \\
        ViT       &  85.52  &  82.81  &  85.45  &  82.82  \\
        Swin &  86.90  &  85.36  &  87.70  &  85.24  \\
        \midrule
        \multicolumn{5}{l}{\emph{Full Tuning}} \\
        \midrule
        ResNet-50 &  94.14  &  90.38  &  93.58  &  91.35  \\
        VGG-16    &  92.41  &  88.84  &  93.09  &  89.62  \\
        Eff.-B1 &  \textbf{95.51}  &  \textbf{93.52}  &  95.27  &  \textbf{94.08}  \\
        ViT       &  95.17  &  92.77  &  93.38  &  92.81  \\
        Swin &  93.79  &  91.96  &  94.47  &  92.77  \\
        \midrule
        \multicolumn{5}{l}{\emph{Metric Learning}} \\
        \midrule
        ResNet-50 (ArcFace)  &  93.45  &  89.48  &  93.28  &  90.70  \\
        ResNet-50 (Center)   &  94.14  &  92.69  &  95.13  &  93.35  \\
        VGG-16 (ArcFace)   &  75.86  &  65.97  &  70.22  &  66.62  \\
        VGG-16 (Center)   &  76.90  &  70.44  &  69.09  &  68.60  \\
        Eff.-B1 (ArcFace) &  93.79  &  91.68  &  94.57  &  92.38  \\
        Eff.-B1 (Center) &  94.48  &  93.21  &  \textbf{95.51}  &  93.67  \\
        ViT (ArcFace)      &  73.79  &  72.24  &  74.79  &  72.14  \\
        ViT (Center)      &  86.90  &  81.56  &  88.34  &  83.45  \\
        Swin (ArcFace)&  71.38  &  68.68  &  69.56  &  67.81  \\
        Swin (Center)&  87.24  &  83.55  &  88.66  &  85.18  \\
        \midrule
        \multicolumn{5}{l}{\emph{Local Features}} \\
        \midrule
        SIFT  &  83.34  &  81.94  &  84.27  &  83.11  \\
        ORB   &  79.92  &  78.81  &  79.04  &  76.57  \\
        \midrule
        \multicolumn{5}{l}{\emph{Integrating Local Features}} \\
        \midrule
        SIFT + ResNet (Full) &  93.45  &  90.18  &  93.67  &  91.16  \\
        \bottomrule
    \end{tabular}
    }
    \caption{Performance comparison on \textsc{PawPrint} dataset.}
    \label{tab:performance_comparison}
\end{table}

\begin{table}[t]
    \centering
    \resizebox{\columnwidth}{!}{%
    \begin{tabular}{lcccc}
        \toprule
        \textbf{Method} & \textbf{Accuracy} & \textbf{Recall} & \textbf{Precision} & \textbf{F1-score} \\
        \midrule
        \multicolumn{5}{l}{\emph{Linear Probing}} \\
        \midrule
        ResNet-50 &  24.77  &  10.39  &  12.00  &  9.64  \\
        VGG-16    &  20.80  &  16.52  &  7.34  &  7.20  \\
        Eff.-B1 &  30.13  &  12.22  &  20.99  &  12.52  \\
        ViT       &  17.56  &  11.81  &  14.25  &  11.00  \\
        Swin &  17.56  &  19.32  &  18.35  &  11.90  \\
        \midrule
        \multicolumn{5}{l}{\emph{Full Tuning}} \\
        \midrule
        ResNet-50 &  32.35  &  \textbf{19.85}  &  18.81  &  12.42  \\
        VGG-16    &  24.58  &  15.05  &  15.60  &  13.31  \\
        Eff.-B1 &  19.69  &  16.99  &  19.70  &  15.16  \\
        ViT       &  12.20  &  15.32  &  12.80  &  10.78  \\
        Swin &  14.70  &  18.57  &  15.26  &  12.08  \\
        \midrule
        \multicolumn{5}{l}{\emph{Metric Learning}} \\
        \midrule
        ResNet-50 (ArcFace)  &  35.77  &  14.49  &  20.76  &  12.58  \\
        ResNet-50 (Center)   &  16.27  &  14.26  &  \textbf{24.09}  &  12.39  \\
        VGG-16 (ArcFace)   &  \textbf{38.45}  &  8.33  &  3.20  &  4.63  \\
        VGG-16 (Center)   &  36.88  &  9.00  &  4.22  &  5.65  \\
        Eff.-B1 (ArcFace)&  21.44  &  16.30  &  19.90  &  15.40  \\
        Eff.-B1 (Center)&  30.41  &  18.37  &  17.29  &  14.49  \\
        ViT (ArcFace)      &  15.80  &  13.77  &  20.33  &  10.56  \\
        ViT (Center)      &  26.80  &  13.09  &  10.35  &  10.83  \\
        Swin (ArcFace)&  35.68  &  13.25  &  14.78  &  11.69  \\
        Swin (Center)&  29.21  &  12.43  &  11.19  &  11.25  \\
        \midrule
        \multicolumn{5}{l}{\emph{Local Features}} \\
        \midrule
        SIFT  &  34.68  &  14.01  &  15.49  &  12.19  \\
        ORB   &  22.64  &  9.40  &  7.97  &  8.07  \\
        \midrule
        \multicolumn{5}{l}{\emph{Integrating Local Features}} \\
        \midrule
        SIFT + ResNet (Full) &  38.36  &  18.11  &  20.80  &  \textbf{16.59}  \\
        \bottomrule
    \end{tabular}
    }
    \caption{Performance comparison on \textsc{PawPrint+} dataset.}
    \label{tab:performance_comparison_PP+}
\end{table}

\section{Animal Identification}
We evaluate a range of methods for individual animal identification on both \textsc{PawPrint} and \textsc{PawPrint+}. 
Our goals are twofold: (1) to benchmark standard deep learning architectures under various training strategies, 
and (2) to examine whether classical local features (SIFT, ORB) remain competitive when data is limited or footprints are highly variable.

\subsection{Implementation Details}

\paragraph{Models and Training.}
We consider five ImageNet-pretrained architectures:
VGG-16, ResNet-50~\cite{he2016deep},
EfficientNet-B1 (Eff.-B1)~\cite{tan2019efficientnet},
Vision Transformer (ViT-B)~\cite{dosovitskiy2020image},
and Swin Transformer (Swin-B)~\cite{liu2021swin}.
Additionally, we include local feature descriptors:
Scale-Invariant Feature Transform (SIFT)~\cite{lowe2004distinctive} 
and Oriented FAST and Rotated BRIEF (ORB)~\cite{rublee2011orb}.
For deep networks, we compare:
(\emph{i}) Linear Probing, 
(\emph{ii}) Full Finetuning (cross-entropy), 
(\emph{iii}) Metric Learning (cross-entropy + ArcFace~\cite{deng2019arcface} or Center Loss~\cite{wen2016discriminative}).
Local features are extracted and then classified via a lightweight two-layer linear head.

\paragraph{Hyperparameters.}
All models are trained for 100 epochs using AdamW with an initial learning rate of 0.001, 
a cosine annealing schedule, weight decay of 0.01, and batch size of 32. 
Images are resized to $224 \times 224$ and augmented by random flips, rotations, color jitter, and Gaussian blur. 
A weighted random sampler addresses class imbalance. 
We report Top-1 accuracy, recall, precision, and F1-score.

\subsection{Quantitative Results}
The quantitative results evaluate various animal identification methods on the \textsc{PawPrint} and \textsc{PawPrint+} datasets.
Performance metrics, including Top-1 accuracy, recall, precision, and F1-score, are detailed in Table~\ref{tab:performance_comparison} for \textsc{PawPrint}  and Table~\ref{tab:performance_comparison_PP+} for \textsc{PawPrint+}.

\subsubsection{Performance in Controlled Environments}

\paragraph{Deep Learning Models.}
Under full finetuning, CNN-based architectures consistently achieve high performance with most exceeding 90\% accuracy.
EfficientNet-B1 achieves the best overall performance (95.51\% accuracy, 94.08\% F1-score), followed closely by ViT at 95.17\% accuracy.
These results confirm that modern neural architectures can effectively learn discriminative features from pawprint images when substrate conditions are consistent.

\paragraph{Local Features.}
Classical feature descriptors demonstrate surprising effectiveness, with SIFT achieving 83.34\% accuracy and 83.11\% F1-score.
While these results trail fully-finetuned deep networks, they validate that hand-crafted descriptors can capture discriminative patterns in footprints under controlled conditions. Moreover, ORB features also perform reasonably well (79.92\% accuracy), albeit with lower overall metrics compared to SIFT.


\subsubsection{Performance in Challenging Conditions}

\paragraph{Challenging Setup for Deep Models.}
Even with full finetuning, model performance declines dramatically in challenging conditions.
The best accuracy among deep networks falls to 32.35\% (ResNet-50), with EfficientNet-B1 achieving only 19.69\% accuracy despite being the top performer on PAWPRINT.
This substantial degradation highlights the difficulty of generalizing across diverse substrates with limited training data.
Interestingly, metric learning approaches show variable effects in this challenging setting.
ResNet-50 with ArcFace achieves the highest accuracy among deep models (35.77\%), while Center Loss variants generally underperform their ArcFace counterparts.
VGG-16 with ArcFace achieves relatively high accuracy (38.45\%) but extremely poor precision (3.20\%), indicating issues with false positive predictions.

\paragraph{Local Features.}
SIFT demonstrates noteworthy robustness under challenging conditions, achieving 34.68\% accuracy without any domain-specific training.
This competitive performance against fully-trained deep networks highlights the advantage of engineered features when facing significant domain shifts.

\paragraph{Family-Related Confusions.}
In \textsc{PawPrint+} we observe substantial errors among two family groups: \emph{Family~1} (Dogs~5, 6) and \emph{Family~2} (Dogs~7, 8, 9). 
Family~1 comprises 110 footprints, but collectively achieves only about 20\% accuracy and over 40\% intra-family misclassifications; 
for Family~2 (188 footprints), overall accuracy remains near 23\%, with one third of the prints wrongly attributed to other family members. 
Given these family groups constitute nearly 28\% of the 1{,}082-image test set, such confusion emphasizes how subtle morphological similarities 
among genetically related animals can lead to frequent misidentification, highlighting the need for more advanced feature-learning strategies.

\subsection{Discussion}
Overall, deep learning networks excel in clear environment (\textsc{PawPrint}) but exhibit notable drops in more varied terrains (\textsc{PawPrint+}). 
Local features can mitigate some of this drop, especially if combined with CNN embeddings. 
These results emphasize the continued relevance of classical descriptors for specialized tasks with limited data and high variability. 
Future work may explore domain adaptation or hybrid pipelines that synergize learned representations with robust local cues, 
potentially improving performance and reducing misclassifications among closely related animals.

\begin{figure}[!t]
  \includegraphics[width=\linewidth]{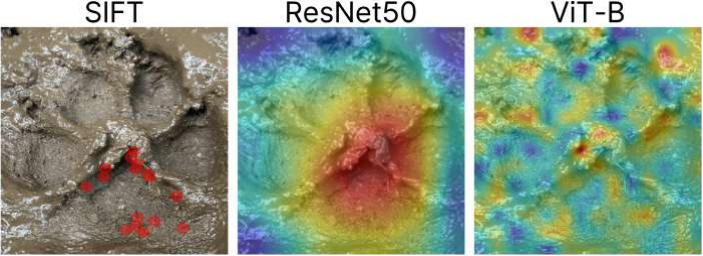}
  \caption{
  This figure compares the key point detection of SIFT, the attention maps of ResNet50 Grad-CAM, and ViT.
SIFT primarily focuses on key landmarks within the footprint, ResNet50 Grad-CAM highlights the overall shape of the footprint, and ViT captures a more global context of the footprint
  }
   \label{fig:animal-pictures}
\end{figure}

\section{Conclusion}\label{sec:conclusion}
In this paper, we introduced the \textsc{PawPrint} and \textsc{PawPrint+} datasets,
the first publicly available resources dedicated to \emph{individual-level} dog and cat footprint identification.
By incorporating diverse real-world terrains,
our datasets enable a more comprehensive evaluation of non-invasive animal identification methods.
We conducted extensive benchmarking using modern deep neural networks and classical local feature techniques,
highlighting how each approach handles varying footprint clarity and environmental complexity.
While deep networks generally excel in controlled conditions, local feature methods showed robustness 
in challenging terrains and limited-data settings.
These findings suggest potential avenues for hybrid strategies that combine learned global representations
with classical local descriptors.
We anticipate that our findings will contribute to the animal identification methodologies,
offering more efficient and ethically sound alternatives for tracking, managing animals and applications in wildlife conservation.


\paragraph{Acknowledgement}
This research was supported by the MSIT (Ministry of Science and ICT), Korea, under the Graduate School of Metaverse Convergence Support Program (IITP-2025-RS-2023-00254129) and the Global Scholars Invitation Program (RS-2024-00459638), both supervised by the IITP (Institute for Information \& Communications Technology Planning \& Evaluation)

\vfill\pagebreak

\bibliographystyle{IEEEbib}
\bibliography{strings,refs}

\begin{thebibliography}{10}

\bibitem{APPA2023pet}
American Pet~Products Association,
\newblock ``{Pet Industry Market Size, Trends \& Ownership Statistics},'' 2023,
\newblock [Online; accessed 02-December-2023].

\bibitem{rodriguez2022trends}
Jeffrey~R Rodriguez, Jon Davis, Samantha Hill, Peter~J Wolf, Sloane~M Hawes, and Kevin~N Morris,
\newblock ``Trends in intake and outcome data from us animal shelters from 2016 to 2020,''
\newblock {\em Frontiers in Veterinary Science}, 2022.

\bibitem{marston2004happens}
Linda~C Marston et~al.,
\newblock ``What happens to shelter dogs? an analysis of data for 1 year from three australian shelters,''
\newblock {\em Journal of Applied Animal Welfare Science}, 2004.

\bibitem{canal2018dogs}
David Canal, Beatriz Mart{\'\i}n, Manuela De~Lucas, and Miguel Ferrer,
\newblock ``Dogs are the main species involved in animal-vehicle collisions in southern spain: Daily, seasonal and spatial analyses of collisions,''
\newblock {\em PLoS One}, vol. 13, no. 9, pp. e0203693, 2018.

\bibitem{bernete2021amelioration}
Eva Bernete~Perdomo, Jorge~E Ara{\~n}a~Padilla, and Siegfried Dewitte,
\newblock ``Amelioration of pet overpopulation and abandonment using control of breeding and sale, and compulsory owner liability insurance,''
\newblock {\em Animals}, vol. 11, no. 2, pp. 524, 2021.

\bibitem{schilling2022review}
Anna-Katarina Schilling, Maria~Vittoria Mazzamuto, and Claudia Romeo,
\newblock ``A review of non-invasive sampling in wildlife disease and health research: What’s new?,''
\newblock {\em Animals}, vol. 12, no. 13, pp. 1719, 2022.

\bibitem{li2018using}
Binbin~V Li, Sky Alibhai, Zoe Jewell, Desheng Li, and Hemin Zhang,
\newblock ``Using footprints to identify and sex giant pandas,''
\newblock {\em Biological Conservation}, vol. 218, pp. 83--90, 2018.

\bibitem{hiby2009tiger}
Lex Hiby et~al.,
\newblock ``A tiger cannot change its stripes: using a three-dimensional model to match images of living tigers and tiger skins,''
\newblock {\em Biology letters}, vol. 5, no. 3, pp. 383--386, 2009.

\bibitem{anderson2010computer}
Carlos~JR Anderson et~al.,
\newblock ``Computer-aided photo-identification system with an application to polar bears based on whisker spot patterns,''
\newblock {\em Journal of Mammalogy}, vol. 91, no. 6, pp. 1350--1359, 2010.

\bibitem{lahiri2011biometric}
Mayank Lahiri et~al.,
\newblock ``Biometric animal databases from field photographs: identification of individual zebra in the wild,''
\newblock in {\em Proceedings of the 1st ACM international conference on multimedia retrieval}, 2011, pp. 1--8.

\bibitem{bae2021dog}
Han~Byeol Bae, Daehyun Pak, and Sangyoun Lee,
\newblock ``Dog nose-print identification using deep neural networks,''
\newblock {\em IEEE Access}, 2021.

\bibitem{vcermak2024wildlifedatasets}
Vojt{\v{e}}ch {\v{C}}erm{\'a}k et~al.,
\newblock ``Wildlifedatasets: An open-source toolkit for animal re-identification,''
\newblock in {\em WACV}, 2024.

\bibitem{shinoda2025petface}
Risa Shinoda and Kaede Shiohara,
\newblock ``Petface: A large-scale dataset and benchmark for animal identification,''
\newblock in {\em ECCV}. Springer, 2025.

\bibitem{alibhai2008footprint}
Sky~K Alibhai, Zoe~C Jewell, and Peter~R Law,
\newblock ``A footprint technique to identify white rhino ceratotherium simum at individual and species levels,''
\newblock {\em Endangered Species Research}, vol. 4, no. 1-2, pp. 205--218, 2008.

\bibitem{kistner2022s}
Frederick Kistner et~al.,
\newblock ``It's otterly confusing: Distinguishing between footprints of three of the four sympatric asian otter species using morphometrics and machine learning,''
\newblock {\em OTTER, Journal of the International Otter Survival Fund}, vol. 2022, 2022.

\bibitem{kistner2023can}
F~Kistner, L~Slaney, and N~Morant,
\newblock ``Can you tell the species by a footprint?-identifying three of the four sympatric southeast asian otter species using computer vision and deep learning,''
\newblock {\em IUCN Otter Spec. Group Bull}, vol. 40, no. 4, pp. 197--210, 2023.

\bibitem{shinoda2024openanimaltracks}
Risa Shinoda and Kaede Shiohara,
\newblock ``Openanimaltracks: A dataset for animal track recognition,''
\newblock {\em arXiv preprint arXiv:2406.09647}, 2024.

\bibitem{girault2022dog}
C~Girault, Nathalie Priymenko, M~Helsly, C~Duranton, and F~Gaunet,
\newblock ``Dog behaviours in veterinary consultations: Part 1. {E}ffect of the owner’s presence or absence,''
\newblock {\em the Veterinary journal}, vol. 280, pp. 105788, 2022.

\bibitem{he2016deep}
Kaiming He et~al.,
\newblock ``Deep residual learning for image recognition,''
\newblock in {\em CVPR}, 2016.

\bibitem{tan2019efficientnet}
Mingxing Tan and Quoc Le,
\newblock ``Efficientnet: Rethinking model scaling for convolutional neural networks,''
\newblock in {\em International conference on machine learning}. PMLR, 2019, pp. 6105--6114.

\bibitem{dosovitskiy2020image}
Alexey Dosovitskiy,
\newblock ``An image is worth 16x16 words: Transformers for image recognition at scale,''
\newblock {\em arXiv preprint arXiv:2010.11929}, 2020.

\bibitem{liu2021swin}
Ze~Liu et~al.,
\newblock ``Swin transformer: Hierarchical vision transformer using shifted windows,''
\newblock in {\em ICCV}, 2021.

\bibitem{lowe2004distinctive}
David~G Lowe,
\newblock ``Distinctive image features from scale-invariant keypoints,''
\newblock {\em International journal of computer vision}, vol. 60, pp. 91--110, 2004.

\bibitem{rublee2011orb}
Ethan Rublee et~al.,
\newblock ``Orb: An efficient alternative to sift or surf,''
\newblock in {\em 2011 International conference on computer vision}. Ieee, 2011.

\bibitem{deng2019arcface}
Jiankang Deng, Jia Guo, Niannan Xue, and Stefanos Zafeiriou,
\newblock ``Arcface: Additive angular margin loss for deep face recognition,''
\newblock in {\em CVPR}, 2019.

\bibitem{wen2016discriminative}
Yandong Wen et~al.,
\newblock ``A discriminative feature learning approach for deep face recognition,''
\newblock in {\em ECCV}. Springer, 2016.

\end{thebibliography}

\end{document}